\documentclass[10pt,twocolumn,letterpaper]{article}

\usepackage[pagenumbers]{wacv} % To force page numbers, e.g. for an arXiv version

\usepackage{graphicx}
\usepackage{amsmath}
\usepackage{amssymb}
\usepackage{booktabs}
\usepackage{multirow}
\usepackage{array}
\usepackage{tabularx}
\usepackage{caption}
\usepackage{subcaption}
\usepackage{tabularray}

\usepackage[accsupp]{axessibility}

\usepackage[pagebackref,breaklinks,colorlinks]{hyperref}

\usepackage[capitalize]{cleveref}
\crefname{section}{Sec.}{Secs.}
\Crefname{section}{Section}{Sections}
\Crefname{table}{Table}{Tables}
\crefname{table}{Tab.}{Tabs.}

\def\wacvPaperID{1650} % *** Enter the WACV Paper ID here
\def\confName{WACV}
\def\confYear{2025}

\begin{document}

%%%%%%%%% TITLE - PLEASE UPDATE
\title{KDC-MAE: Knowledge Distilled Contrastive Mask Auto-Encoder}

\author{Maheswar Bora, Saurabh Atreya, Aritra Mukherjee, Abhijit Das\\
Birla Institute of Technology and Science, Pilani – Hyderabad Campus \\
Secunderabad, Telangana 500078\\
{\tt\small abhijit.das@hyderabad.bits-pilani.ac.in}
% For a paper whose authors are all at the same institution,
% omit the following lines up until the closing ``}''.
% Additional authors and addresses can be added with ``\and'',
% just like the second author.
% To save space, use either the email address or home page, not both
}
\maketitle
%%%%%%%%% ABSTRACT
\begin{abstract}
In this work, we attempted to extend the thought and showcase a way forward for the Self-supervised Learning (SSL) learning paradigm by combining contrastive learning, self-distillation (knowledge distillation) and masked data modelling, the three major SSL frameworks, to learn a joint and coordinated representation. The proposed technique of SSL learns by the collaborative power of different learning objectives of SSL. Hence to jointly learn the different SSL objectives we proposed a new SSL architecture KDC-MAE, a complementary masking strategy to learn the modular correspondence, and a weighted way to combine them coordinately. Experimental results conclude that the contrastive masking correspondence along with the KD learning objective has lent a hand to performing better learning for multiple modalities over multiple tasks.   %As  a  result,  our  fullyself-supervised pre-trained CAV-MAE achieves a new SOTA accuracy of 65.9% onVGGsound, and is comparable with the previous best supervised pre-trained modelon AudioSet in the audio-visual event classification task.   
%Code and pre-trained models are at \textcolor{red}{https://github.com/MBora/Sweet-point-SSL} 
\end{abstract}

%%%%%%%%% BODY TEXT
\section{Introduction}

\label{sec:introduction}
A natural tendency of human beings is to learn about the environment by themselves or be self-taught''. Since our childhood, we start learning via supervised training. Slowly we start analyzing and learning things by ourselves or start getting ``self-taught''. There are several aspects by which we get self-taught and a combination of them builds our life. 

Even the learning representation community adopted the self-taught~\cite{raina2007self} as SSL. %SSL is treated separately...
Learning without training typically involves recalling the data, grouping similar things and unrelated separately for learning the abstract knowledge from data.
%Incidentally, while self-learning \textit{i.e.} learning without supervision, we as a human being try to analyse things. We try to imagine something from its related partial visual description, or group things based on their similarity or disparity, learning from similar scenarios to knowledge exchange between different forms of the same scenario. 
%In fact, SSL learns using learning objectives, which is very similar to how humans learn by analysis while self-learning.
Common SSL learning objectives are \textit{pretext task}~\cite{lee2021predicting}, \textit{contrastive learning}~\cite{jaiswal2020survey}, and \textit{teacher-student} methods~\cite{bucci2021self, schiappa2023self, wang2021knowledge}.
%The popular learning objectives for SSL had been pretext learning, reconstruction, contrastive learning and masked data modelling.\\
For pretext learning on images one popular task was colorization~\cite{iizuka2016let, larsson2016learning, larsson2017colorization} and jigsaw puzzle solving~\cite{noroozi2016unsupervised, kim2018learning, noroozi2018boosting, carlucci2019domain, wei2019iterative, salehi2020puzzle, ren2023masked}. In the past, the jigsaw puzzle rearrangement has been used both on spatial and temporal domains for learning better video representations~\cite{wang2022video}. 
Various works~\cite{srivastava2015unsupervised, vondrick2016generating, tulyakov2018mocogan, wang2022bevt} used video reconstruction as learning objectives by disentangling motion and content, using scene dynamics and transformers. 
Super-resolution is gaining popularity as a learning objective for scale-independent generalized learning~\cite{xia2023structured}.
%Recently super-resolution has also been a learning objective to learn scale-independent visual features in a generalized way~\cite{xia2023structured}.\\

\begin{figure*}[h!]
    \centering
    \includegraphics[width=14cm]{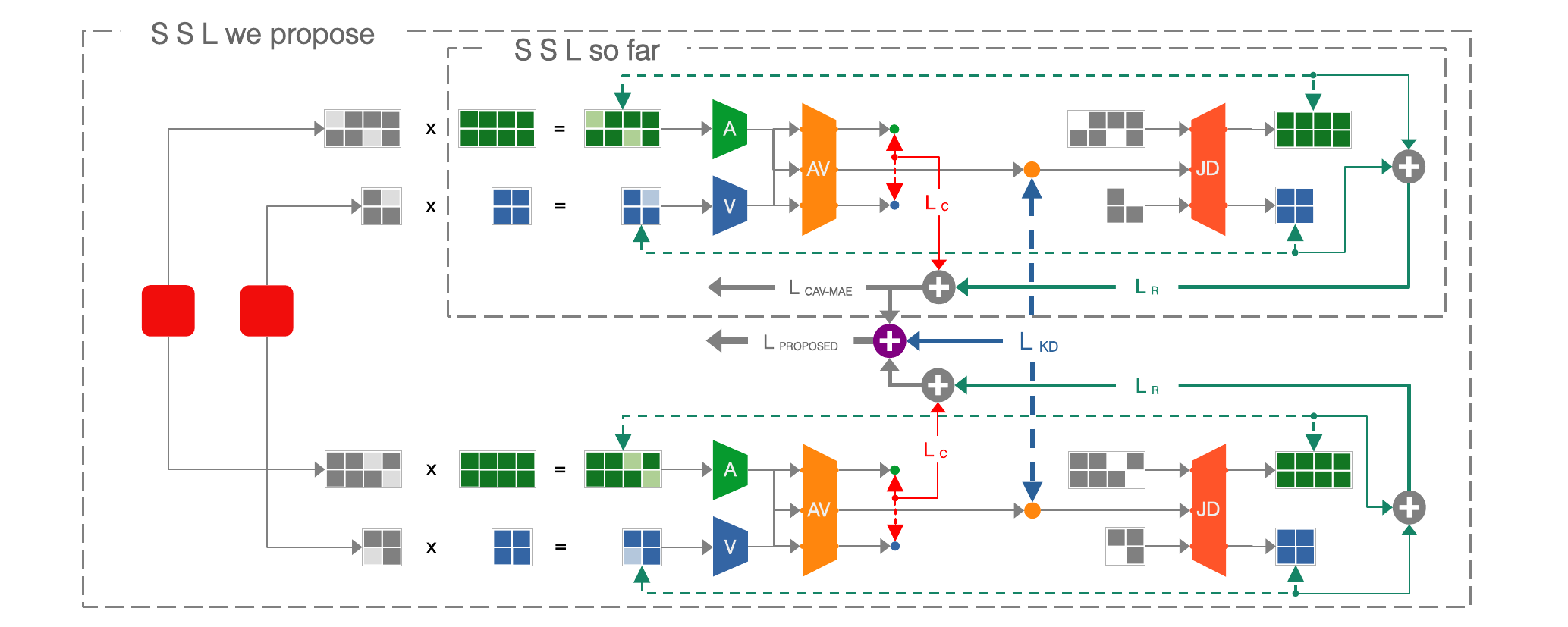}
    \caption{Proposed improvement on existing SSL by complementary mask and self-distillation. The model uses shared weights with two masked versions of the same audio-video pair passed through the encoder, generating separate joint latent embeddings. KL divergence loss aligns these embeddings, followed by a joint decoder that splits audio and video, with contrastive loss applied to the latent embeddings.}
    \label{fig:graphic_abstract}
\end{figure*}

Contrastive learning is a widely used learning objective. A comprehensive survey~\cite{feichtenhofer2021large} shows that approaches can be classified into three categories. First, Binary classification of ``good pair'' and ``bad pair'' using Binary Cross Entropy (BCE) loss~\cite{knights2021temporally, yao2021seco}. Second, SSL can be done with discriminators, which is the fuzzy version of the binary classification~\cite{tao2020self, wang2021enhancing}. Third, the most popular approach in SSL is to use Noise-Contrastive Estimation (NCE) loss~\cite{gutmann2010noise} to pretrain the model by pulling similar data pairs closer and pushing dissimilar data pairs further. Though originally devised for Natural language Processing (NLP) tasks~\cite{mnih2013learning}, the concept is now widely used in video~\cite{han2020self, lorre2020temporal, qian2021spatiotemporal, yuan2022contextualized, kordopatis2023self}, images~\cite{chaitanya2020contrastive, nguyen2023joint, denize2023similarity} and audio~\cite{liu2022audio}.

In recent years since Masked Autoencoders (MAE) was proposed for vision tasks~\cite{he2022masked}, the domain has been heavily explored by researchers resulting in a huge amount of research outcomes in recent years. Due to huge interest in this area, comprehensive surveys~\cite{zhang2022survey,zhang2023survey, zhou2023masked} were made in recent times. Subsequently, video-MAE~\cite{tong2022videomae,feichtenhofer2022masked, wei2022masked}, Audio-visual MAE\cite{georgescu2023audiovisual} and CAV-MAE~\cite{gong2022contrastive} are proposed on this direction of research. In recent time-motion-guided masking~\cite{yang2022self, huang2023mgmae} and dual masking~\cite{wang2023videomae} have been explored in this regard.
It is observed that the selection of a masking policy has a profound impact on the learnability of models. Knowledge distillation also plays an important role in self-distillation between instances with different masking policies~\cite{chen2022sdae, zhang2022mask}. It is also observed that knowledge distillation can provide a significant boost to SSL~\cite{zhang2018deep, grill2020bootstrap, caron2021emerging, oquab2023dinov2}. 

The aforementioned learning objectives of SSL are well explored individually and found to perform nonuniformly for different scenarios when they are employed individually. It is expected that they will work better if they learn jointly to find mutual correspondence. 
%The processes are better known as masked autoencoding, contrastive learning and self-knowledge distillation.
%SSL is one of the paradigms of machine learning that holds a lot of promise for training models for real-life applications. Due to the web-scale amount of data, without annotation, that SSL can handle, it constructs a promising ground for myriads of downstream tasks by building a generic learner. 
A recent work \cite{gong2022contrastive} attempted to unify the concepts of masked autoencoder for the reconstruction of video frames and contrastive learning between video and audio tokens to learn better multimodal representations. 
%The reason for concentrating on CAVMAE is elaborated next. But before that we must discuss about the evolution of SSL till now.\\
%A survey work on application of SSL in Convolutional Neural Network (CCN) based models for image and video modality~\cite{jing2020self} shows that the major aspects of SSL is either reconstruction through an autoencoder or contrastive learning through a Siamese model or a combination of both. The reconstruction task can have pretexts like image colorization~\cite{zhang2016colorful} and inpainting~\cite{pathak2016context} and the contrastive task can have pretext like solving jigsaw puzzles~\cite{noroozi2016unsupervised}. With the introduction of Vision Transformers(ViT)~\cite{dosovitskiy2020image}, researchers started exploring SSL on ViT.\\
%We observed a culmination of the major aspects in CAV-MAE which efficiently combined masked data modelling and contrastive learning. CAV-MAE enhanced audio classification by applying SSL on an MAE during pretraining along with partial joint learning between the modalities. Thus employing the two sub paradigms of SSL, \textit{\textit{i.e.}} training by reconstruction and contrastive loss, it elevated the efficacy of audio-visual learning hugely.

%We found the claim interesting that masked data modelling and contrastive learning are complementary and thus we started a deeper investigation into the matter, grounding our experiments on the CAVMAE vanilla architecture. 
\textit{The reconstruction task of CAV-MAE forces its representation to encode the input information in the fusion and the constructive task helps to find the explicit audio-visual correspondence objective but lacks in finding the modular correspondence. This motivates us to propose Knowledge Distilled Contrastive Mask Auto-Encoder (KDC-MAE), an optimized manner to use modular knowledge distillation along with masked data modelling, and inter-modal contrastive learning to learn joint multimodal correspondence of audio and video for unified learning representation}. We proposed to use dual complementary masks as input to dual head mask-autoencoder in a weight-sharing fashion to add modular correspondence. Further, to foster the modular correspondence, Kullback–Leibler divergence (KL) of the representation of the dual mask of each modality is computed in the encoding space and propagated as a loss (See Fig.~\ref{fig:graphic_abstract}). 

From our experiments, we conclude that these learning objectives can boost learning in an optimal combination. We demonstrate our findings through various ablation studies at the pretraining level. To support our claim we utilized the pre-trained models in further several downstream tasks over different modalities and their combinations, on which we obtained better results.
In summary, our contributions are: 
% \vspace{-3mm}
\begin{itemize}
\item To propose that three paradigms of SSL namely contrastive learning, masked data modelling, and knowledge distillation objectives are not complementary for unified joint modelling.
\item Introduction of complementary masking strategy for finding the modular correspondence of the modalities. 
% \vspace{-3mm}
\item Introduction of static regularizers as weights for equating the various losses of the different tasks of SSL for joint and collaborative learning to generalise the contextual information extraction capability of the encoder.
% \vspace{-4mm}

%% \vspace{-1mm}
%\item Rigorous analysis of various combinations of learning objectives for SSL and coming up with an optimal ensemble of them.

\end{itemize}

\section{Proposed Methodology}
% \vspace{-2mm}
In our proposed work we focused on optimization of the ensemble of the SSL, few questions arose for the training approach: (a) Investigating whether reconstruction from masked input, contrastive learning and knowledge distillation is complementary; (b) If they are not complementary, in what way can they be best combined? (b) How can we incorporate mutual learning as a form of self distillation~\cite{zhang2018deep}, with different unmasked tokens as input to weight shared instances of the encoder, to train it better? (d) How can we make the SSL learning task tougher by experimenting with more masking strategies in order to make the encoder learn better?

% \begin{figure*}[h]
%     \centering
%     \includegraphics[scale=0.5]{Index.pdf}
%     % \vspace{-4mm}
%     \caption{Symbol Index of building blocks of the network and arrangements that we experimented with}
    
%     \label{fig:index}
% \end{figure*}
% % \vspace{-1mm}
% \begin{figure*}[h]
%     \centering
%     \includegraphics[scale=0.5]{CAVMAE_vanilla.pdf}
%     % \vspace{-4mm}
%     \caption{The vanilla CAV-MAE~\cite{gong2022contrastive} model, note that the normalization layers of audio, video and joint tokens are separate. Also refer to Fig.~\ref{fig:index} for the index of meanings of the different components of the figure.}
    
%     \label{fig:cavmae_vanilla}
% \end{figure*}

% % \vspace{-2mm}

% \begin{figure*}[h]
%     \centering
%     \includegraphics[scale=0.5]{comp_mask.pdf}
%     % \vspace{-7mm}
%     \caption{The complementary mask generator working principle, note two video masks and audio masks are complementary but there is no relation between any two audio and video masks}
    
%     \label{fig:comp_mask}
% \end{figure*}

%% \vspace{
\subsection{Preliminaries}
% \vspace{-2mm}
%assumptions, building blocks, notations
We follow a similar technique for pre-processing and tokenization as in MAE \cite{dosovitskiy2020image} which was further adopted by CAV-MAE ~\cite{gong2022contrastive}. Hence, we take on audio-visual data by considering a frame of the $10$sec video clip as the visual input and the Mel Spectrogram of the entire audio as an image, as the audio input. The encoder architecture for both modalities is the same and the decoder is common. The heart of the encoder is based on transformer layers~\cite{vaswani2017attention}. 
%which in turn contains multi-headed self-attention (MSA), layer normalization (LN), and multilayer perceptron (MLP) blocks with residual connections. 
Both the audio encoder(AE) and video encoder(VE) are composed of $11$ transformer layers and the joint audio-video encoder(AVE) contains $1$ layer. 
%In easy terms, the modality-specific information is projected for self-attention eleven times what is done in joint modality mode. 
The total $12$ layer encoder is based on the ViT base model~\cite{dosovitskiy2020image}. Thus its input is also $16 \times 16$ patches, termed as unmasked tokens. The video input frames are scaled-center-cropped $224 \times 224$ thus resulting in $196$ patches. The audio spectrogram is of size $128 \times 1024$ (features $\times$ time) and thus it results in $512$ $16 \times 16$ patches. The masking ratio was kept at $75\%$ and thus $49$ video patches and $128$ audio patches were the input for the audio and the video encoder. 
After passing through the $11$ layers of both the embeddings are concatenated and passed through the single layer of the audio-video encoder. Three sets of embeddings, \textit{i.e.} the video, audio and audio video are passed in three consecutive passes to keep the time complexity in check. The audio and video embeddings of the \textit{i}th sample ($a_i,v_i$) are subjected to meanpool operation to compute ($c^v_i,c^a_j)$ for further calculation of contrastive loss $\mathcal{L}_\mathrm{c}$. The concatenated embeddings ($\mathbf{x^\prime}$) are passed through the joint decoder (JD) which has $16$ attention heads and has $8$ layers. Both audio and video embeddings pass through common $7$ layers of JD and the last modality-specific layer for final patch reconstruction. The masked tokens are passed as trainable masked tokens, $16 \times 16$ patches attached to the computation graph, during the process. Each reconstructed token gets into its proper place due to modality-specific sinusoidal position embedding. The idea of using zero-valued patches is best explained in~\cite{cao2022understand}. The reconstruction loss $\mathcal{L}_\mathrm{r}$ is computed between the masked patches ($75\%$ of the video frame and the spectrogram). The equations for $\mathcal{L}_\mathrm{c}$ and $\mathcal{L}_\mathrm{r}$ are as follows:
% \vspace{-4mm}
\begin{equation}
\label{equ:contrastive}
\mathcal{L}_\mathrm{c} = - \frac{1}{N} \sum_{i=1}^N {\rm log}  \left[ \frac{ {\rm exp} (s_{i,i}/\tau)}{\sum_{k \neq i} {\rm exp} (s_{i,k}/\tau) + {\rm exp} (s_{i,i}/\tau)} \right] 
\end{equation}
% \vspace{-2mm}
where $s_{i,j} = \|c^v_i\|^T\|c^a_j\|$ and $\tau$ is a temperature parameter. 
% \vspace{-1mm}
\begin{equation}
    \hat{\mathbf{a}}_i, \hat{\mathbf{v}}_i = \mathrm{D}(\mathbf{x^\prime} + [\mathbf{E_a^\prime}, \mathbf{E_v^\prime}] + [\mathbf{E^p_a}^\prime, \mathbf{E^p_v}^\prime]) 
\end{equation}
% \vspace{-2mm}
\begin{equation} \small
    \mathcal{L}_\mathrm{r} = \frac{1}{N} \sum_{i=1}^N
    \left[\frac{\sum(\hat{\mathbf{a}}_i^{\mu}-\mathrm{norm}(\mathbf{a}_i^{\mu}))^2}{|\mathbf{a}_i^{\mu}|} 
    + \frac{\sum(\hat{\mathbf{v}}_i^{\mu}-\mathrm{norm}(\mathbf{v}_i^{\mu}))^2}{|\mathbf{v}_i^{\mu}|}\right]
\end{equation}
where $N$ is the mini-batch size; $\mathbf{a}^{\mu}$, $\mathbf{v}^{\mu}$, $\hat{\mathbf{a}}^{\mu}$, $\hat{\mathbf{v}}^{\mu}$ denote the original and predicted masked patches; $|\mathbf{a}^{\mu}|$ and $|\mathbf{v}^{\mu}|$ denote the number of masked audio and visual patches, $\mathrm{D()}$ is the joint decoder (JD), $\mathbf{E_a^\prime}$ and $\mathbf{E_v^\prime}$ are modality type embedding (for $\mathrm{D()}$ to recognize audio and video tokens), $\mathbf{E^p_a}^\prime$ and $\mathbf{E^p_v}^\prime$ are modality specific 2D sinusoidal embeddings, respectively. The final loss is computed as $\mathcal{L}_\mathrm{total} = \mathcal{L}_\mathrm{r} + \lambda_c\cdot\mathcal{L}_\mathrm{c}$ where $\lambda_c$ is a regulating parameter empirically chosen as $0.01$.

% % % \vspace{-4mm}
\subsection{Complementary mask: why, what and how?}
% \vspace{-2mm}
% % % \vspace{-2mm}
There are a few important things that we observed revisiting the closely related previous works and the reason why we proposed complementary mask:
%% % \vspace{-2mm}
%\begin{itemize}

\noindent\textbf{$\mathcal{L}_\mathrm{c}$} only does positive pair mining, \textit{i.e.} pull together corresponding audio and video. Thus there is no way to find the correspondence between the modality itself, which is important as we are heavily sampling both the audio and video before encoding. \textit{This motivates us to use self-distillation between modality}.

\noindent Due to the nature of implementation, the JD is unaware of the contrastive learning effect. This measure is necessary as the decoder will be discarded for downstream tasks and thus its only concern should be how to reconstruct the embeddings to the masked input patches, the only possible annotation available for for SSL. \textit{Thus self distillation should be done in the encoded space like constructive task.}

\noindent\textbf{The masking strategy} is random and thus there is no implied correspondence between the video and audio tokens, that are passing through VE and AE respectively. As it is impossible to devise any temporal correlation between random spatial patches of a single video frame and the total audio spectrogram, thus self-distillation between video and audio to find the modular correspondence directly is impossible in the present form of data representation. However, distillation between individual modalities of audio and video and the joint embeddings makes sense as the couples belong to the same modality/category. %\textit{So the distillation should be toughest (a proven prerequisite of SSL on MAE}~\cite{he2022masked}\textit{) for best learning and it may be ensured by subjecting
Thus to materialise the same the duel head MAE is employed with non-overlapping or overlapping complementary patches as input to each heads.
%\end{itemize}

The concept of the complementary mask is simply the generation of two masks so that $M_1$ and $M_2$ are generated as sets of unmasked tokens in such a way that the intersection of them is a null set i.e $M_1 \cap M_2 = \phi$. The concept is pictorially described in Fig.~\ref{fig:approach}. It should be noted that $n$ complementary random masks are only possible for a masking ratio $m_r \geq (n/N)$ where $N$ is the total number of input patches. In our case, $n=2$ thus $m_r \geq 0.5$ will suffice. As the masking ratio is $0.75$ thus there was no need to change the masking ratio.

At first, a normal random mask is generated with masking ratio $m_r$. Let the set of all tokens for audio modality be $U_a=512$. Thus if the first mask $M_1^a$ is considered as a set then the residue tokens left for consideration in $M_2^a$ is $U_a - M_1^a$, thus the set $M_2^a$ is randomly chosen from that residual set so that $|M_1^a| = |M_2^a| = m_r\cdot |U_a|$ and $M_1^a \cap M_2^a = \phi$. Similar masks $M_1^v$ and $M_2^v$ are generated for the video modality from $U_v$ where $|U_v|=196$. It should be noted that there is no relation imposed between the audio and video masks while generating the complementary mask.

The motivation behind complementary patches is making the \textit{embedding mask agnostic to find modal correspondence}, \textit{i.e.} if the embeddings are treated as probability distributions, they should be closer irrespective of the input tokens, provided that they are from the same data point. After all a masked encoder is supposed to be a function $F(x)$ where $x$ is a subset of the input $X$ in the form of randomly chosen patches. Thus if two subsets $x_1$ and $x_2$ are there such as $x_1 \subset X, x_2 \subset X, x_1 \cap x_2=\phi$  are there, $F()$ should be trained in such a way that $F(x_1) \simeq F(x_2)$. As $X$ will be reconstructed from $x_1$ and $x_2$ by a single decoder which is shallower than the encoder, it is the task of the encoder to incorporate the entire information of all patches in the unmasked patches. Thus the best encoder is one that can generate similar input to the decoder irrespective of the input subset, even if the two subsets have no overlap at all. 
% \vspace{-2mm}
%\subsection{Data sampling and masking strategy}
\subsection{Learning strategy of the proposed SSL-model}
 % \vspace{-2mm}
Gong et al.~\cite{gong2022contrastive} already proved different learning strategies of SSL are not complementary. Their work demonstrated that a combination of contrastive and reconstruction loss works well towards learning a generic multimodal encoder for SSL. However, they have not explored all combinations of SSL strategies to optimize their combination. Hence, we introduced in this context self-distillation along with contrastive-mask encoding. The idea of self-distillations works like contrastive learning here but the contrast is between two forms of the same input across all modalities and their combinations, rather than just two modalities of the same input. In the reconstruction loss, the accuracy of the masked token reconstruction is checked and is trained to the patches of the masked input patches with the help of information provided in the JD, by the encoded audio-visual embeddings. Our goal is to force the embeddings to preserve better information so that it can provide that to the trainable masked token in the JD, in turn improving the power of the encoder. 

%\subsection{Self distillation through linear symmetric KL divergence (Jeffreys divergence)}
DINO is a popular model to start from when it comes to self-distillation with no labels~\cite{caron2021emerging}. But at any one point in time, it is unidirectional in nature. An older approach called Deep Mutual Learning(DML)~\cite{zhang2018deep} appeared to be a better candidate for our consideration after initial experiments. We tuned the nature of DML by redesigning the training methodology with masked data modelling in mind. For that, we wagered our approach on a symmetric form of KL divergence.
The Kullback-Leibler (KL) distance (discrete version) is defined as:
 % \vspace{-2mm}
\begin{equation}
    \mathcal{D}(p_1||p_2) = \sum_{i=0}^{n-1}(p_1(i)log\frac{p_1(i)}{p_2(i)})
\end{equation}
% \vspace{-1mm}
where $p_1$ and $p_2$ are two probability distributions of $n$ distinct samples. In terms of latent embedding the $n$ translates to the dimension size. Though KL divergence provides a very good log-likelihood measure of two distributions, the major caveat is its asymmetric property \textit{i.e.} $\mathcal{D}(p_1||p_2) \neq \mathcal{D}(p_2||p_1)$. Also, the inputs must be a probability distribution or $\sum_{i=0}^{n-1}p(i) = 1$ and $p(i)\geq 0 \, \forall \, i$ which may not be the case for an embedding.

The embeddings between which the KL divergence is measured are the mean pool for all audio and video tokens that will be subjected to pass to JD (without the positional encoding). In our case we meanpool $\mathbf{x^\prime}$ to $\mathbf{\chi}$ where $n=|\mathbf{\chi}|=768$ as the patches have dimension $16 \times 16$ with three channels. To get the embedding for KL divergence we first find the minimum value $\mathbf{\chi}_{min}$ in the embedding vector ($\mathbf{\chi}$) and then sum or subtract that to the vector, \textit{i.e.} $\mathbf{\chi} =\, <\mathbf{\chi}(i) (+/-)|\mathbf{\chi}_{min}|, \, \forall \, i>$, depending on whether $\mathbf{\chi}_{min}$ is negative or positive, respectively. This is done to make sure $\mathbf{\chi}(i) \geq 0 \, \forall i$. 

Then $\mathbf{\chi}$ is converted to a probability distribution by simple linear normalization \textit{i.e.} $p = (\mathbf{\chi}(i)/\sum_{j=0}^{n-1}\mathbf{\chi}(j)), \forall \, i$. 
Something as common as softmax is avoided as that is tuned for classification problems, which exaggerates the divergence of the distribution for the sake of clarity. Our purpose is not classification but a near-accurate projection of $\mathbf{\chi}$ into a probability distribution $p$ so that KL-divergence can be applied.\\
Now to eliminate the asymmetric effect of KL divergence, we use the self-distillation loss ($\mathcal{L}_{kd}$) formula:
% \vspace{-1mm}
\begin{equation}
    \mathcal{L}_{kd} (p_1,p_2)= \frac{\mathcal{D}(p_1||p_2) + \mathcal{D}(p_2||p_1)}{2}
\end{equation}
% \vspace{-1mm}
Where $p_1$ and $p_2$ are the probability distribution projection of the origin corrected mean-pooled audio-video embeddings $\mathbf{\chi}_{1}$ and $\mathbf{\chi}_{2}$ coming from the parallel streams (refer Fig.~\ref{fig:our_approach}) with input masks $M_1$ and $M_2$, respectively. \\
If we consider $\mathbf{x^\prime}$ to be the probability distribution in a translated latent space $\mathbf{L^\prime} \in \Re^{n}$ then a properly learnt encoder $F(x)$ should compute $\mathbf{x^\prime_1} = F(x_1)$ and $\mathbf{x^\prime_2}=F(x_2)$ so that $\mathbf{x^\prime_1} \simeq \mathbf{x^\prime_2}$. For MAE, the decoder $D()$ is getting trained such as $D(\mathbf{x^\prime_1}) = D(\mathbf{x^\prime_2}) = X$. As $D()$ is to be discarded, only the condition $\mathbf{x^\prime_1} \simeq \mathbf{x^\prime_2}$ can ensure such an outcome. Thus we chose to propagate the $\mathcal{L}_{kd}$ only through $F()$, to make it powerful without being dependent on $D()$.  As it is proven~\cite{anderson2004model} that model selection can be best measured with KL divergence from the ground truth distribution, in our case, due to lack of ground truth in $\mathbf{L^\prime}$, mutual ground truth is the only option left. The idea is that whatever may be the true latent representation, the encoder should be able to approximate a distribution near that, irrespective of input. MAE~\cite{he2022masked} has already proved that random masking solves this purpose, thus making the encoder learn better. Hence, complementary inputs will make this divergence, as at the encoder level there is no option for information exchange between the two sets of complementary tokens. 
% \vspace{-1mm}
\subsection{Combining all the losses and learning objectives}
\label{sec:combining_losses}
% \vspace{-3mm}
We have discussed about three losses, this is how to translate to different learning objectives. Contrastive loss $\mathcal{L}_\mathrm{c}$ translates to the learning objective ``audio and video of the same datapoint should be closer''. Reconstruction loss $\mathcal{L}_\mathrm{r}$ translates to ``even with masked input, the deep encoder should extract enough information to use a shallow decoder for reconstructing the total input''. Self Distillation loss $\mathcal{L}_\mathrm{kd}$ translates to ``whatever may be the mask, the latent audio-video representation of a datapoint should not vary.''

A scalar factor $\lambda_{c}=0.01$ for the contrastive loss $\mathcal{L}_\mathrm{c}$ and $\lambda_{kd}=10$ for self-distillation loss $\mathcal{L}_\mathrm{kd}$ was used before adding to the reconstruction loss $\mathcal{L}_\mathrm{r}$. Due to the nature of implementation, during backward pass  $\mathcal{L}_\mathrm{r}$ affects the modality-specific encoders, the joint encoder and the joint decoder but the contrastive loss $\mathcal{L}_\mathrm{c}$ and $\mathcal{L}_\mathrm{kd} $only effects later two. This is due to the way the graph updates weights based on the local gradient computed in the forward pass and the downstream gradient computed in the backward pass.
% % \vspace{-7mm}

% \begin{figure*}[h]
%     \centering
%     \includegraphics[width=140mm]{our_approach.pdf}
%     % \vspace{-1em}
%     \caption{The proposed learning objective ensemble technique over vanilla CAV-MAE. }
%     % \vspace{-1em}
%     \label{fig:our_approach}
% \end{figure*}

\begin{figure*}[h!]
    \centering
    
    \begin{subfigure}[b]{\textwidth}
        \centering
        \includegraphics[width=120mm]{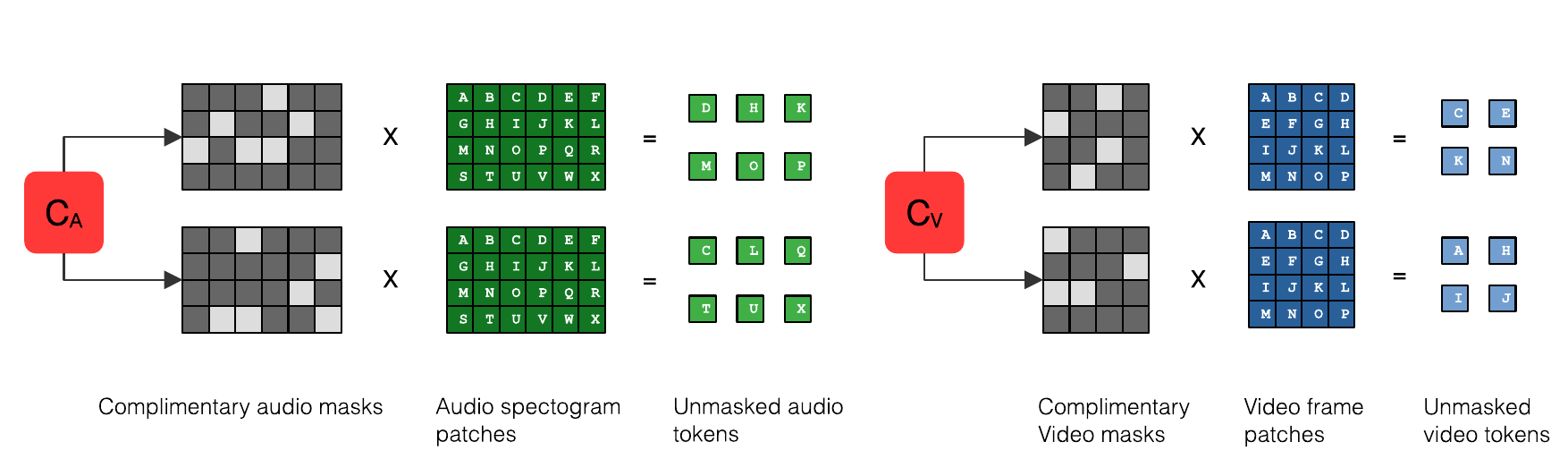}
        \caption{}
        %\label{fig:sub1}
    \end{subfigure}
    \begin{subfigure}[b]{\textwidth}
        \centering
        \includegraphics[width=\linewidth]{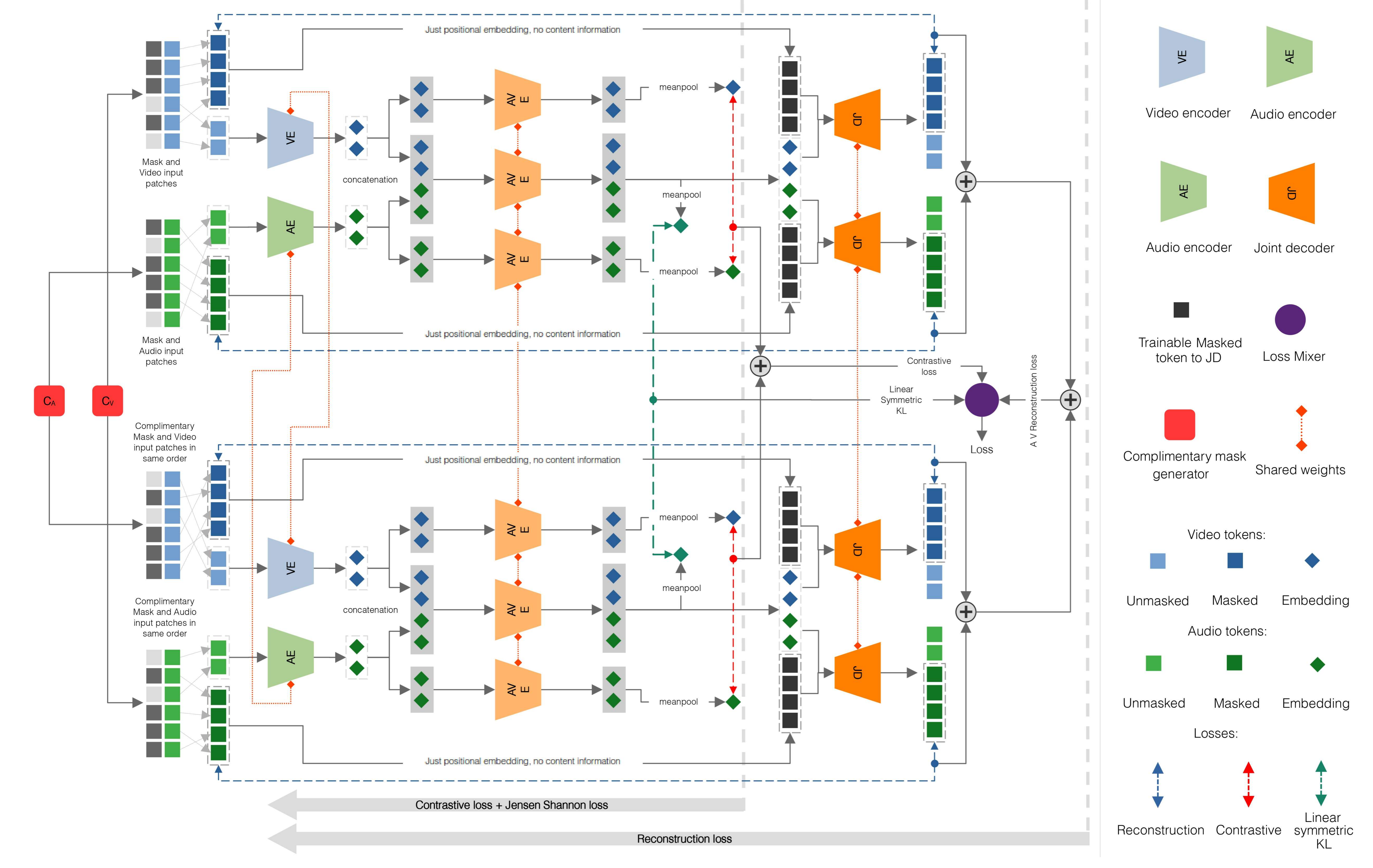}
        \caption{}
        %\label{fig:sub2}
    
    \end{subfigure}
 \caption{(a) The complementary mask generation. Note: Two video masks and audio masks are complementary but there is no relation between any two audio and video masks, (b) The proposed architecture of KDC-MAE: (Symbol index on the right)} 
\label{fig:approach}
\end{figure*}

The self-distillation loss $\mathcal{L}_\mathrm{kd}$ works best for two mutual teacher/student (stream) configurations. For more parallel streams, the performance degrades as proven by experiments. If more than two streams are used then $\mathcal{L}_\mathrm{kd}$ is taken as the mean of all three steams \textit{i.e.}  $\mathcal{L}_\mathrm{kd} (p_1,p_2, p_3) = (\mathcal{L}_\mathrm{kd} (p_1,p_2) + \mathcal{L}_\mathrm{kd} (p_1,p_3) + \mathcal{L}_\mathrm{kd} (p_2,p_3))/3$. In such a configuration the the divergence between two streams affects the third stream, to which it has no contribution. Though the higher number of student cohorts of the parallel stream is supposed to enhance accuracy~\cite{zhang2018deep} the trend reverses due to too much exposure to data as 3 complementary masks expose $75\%$ of the data, which overfits the model. Thus it is proven that a complementary masking strategy is not the holy grail when it comes to training an MAE. 
% % \vspace{-2mm}

%\subsection{Formulating Losses for the different paradigm of SSL}
%what are the losses

%How they are calculated

%How they are formulated

%how the losses are propagating

%How losses justify the proposed SSL
%\subsection{KD}

%\label{sec:proposed_methodology}
%The proposed work tries to build upon the existing technique of CAV-MAE to make it more robust by essentially two techniques:
%\begin{itemize}
%    \item Dynamic weights for losses of reconstruction of both audio and video components, contrastive loss for learning cross context between video and audio and loss between latent representations of encoded and decoded audio and video.
%    \item Knowledge distillation from the leading to lagging encoder in the form of selective loss backpropagation through a novel loss amplification technique. 
%\end{itemize}

%\subsection{Dynamic weight for losses}

%\subsection{Knowledge distillation through selective loss backprop}
% \vspace{-2mm}
\section{Experimental results}
% \vspace{-3mm}
We performed extensive experiments and ablation studies with different learning objectives combined in different ways. Our final configuration is a result of incremental change in various key factors and those results are presented in an ordered fashion to elaborate our path of analysis. We have used $4$ datasets for SSL pretraining and modality-specific finetuning. The downstream task in finetuning has been only classification across different modalities of data.  
% % \vspace{-1em}
\subsection{Datasets and hardware}
% \vspace{-2mm}
The major two datasets that we worked with are AudioSet~\cite{gemmeke2017audio} and VGGsound~\cite{chen2020VGGsound} for training and testing the methodology on audio-visual modality. We have used the Kinetics~\cite{kay2017Kinetics} dataset for video-only action recognition. We have not experimented separately on image modality as the video frame reconstruction is the same task. \\
%For NLP tasks the used dataset is 
%The A G News Classification~\cite{zhang2015character} dataset was used for experiments on NLP tasks.
%C100~\cite{conneau2019unsupervised, wenzek2019ccnet} on NLP tasks. A brief description of each of them are as follows:
\textbf{AudioSet}~\cite{gemmeke2017audio}, is a collection of audio label annotated YouTube clips of $10$ seconds duration. Due to the unavailability of many videos, the dataset on which \textit{we based our experiments has $1472186$ training videos and $7486$ test videos (for label classification)}. The total number of classes is $527$ that are not distributed in a balanced fashion. Our use of AudioSet was mainly in SSL pretraining. A balanced subset called AudioSet $20$k with $8227$ datapoints is used for finetuning tasks. Like the full Audioset, many videos are missing from the original list, in the subset also.\\
\textbf{VGGsound}~\cite{chen2020VGGsound}, is a also collection of audio label annotated YouTube clips of $10$ seconds duration like AudioSet. For the same reason as AudioSet, we were able to experiment with $164003$ training videos and $13560$ test videos. The total number of classes is $300$ and the samples are not balanced over classes.\\
\textbf{Kinetics-400}~\cite{kay2017Kinetics} is also a huge collection of $10$ second YouTube clips where only the video modality is used for human action recognition. We have used $230$k training videos and $19$k validation and testing videos.\\
\textbf{ILSVRC2017}~\cite{ILSVRC15} is a subset of the ImageNet dataset consisting of 1,281,167 training images, 50,000 validation images and 100,000 test images, all belonging to one of 1000 classes.

For all the dataset numbers may vary from the officially released number due to the aforementioned reasons of working with a collection of YouTube videos. 
    %\item \textbf{ILSVRC}~\cite{deng2009imagenet} dataset contains $1.3$m training, $50$k validation and $100$k test images primarily annotated for image classification into $1$k object classes. We have used it primarily to fine tune the encoder for image classification tasks. 
    %\item \textbf{CC100}, is a corpus that was created to train Microsoft's XLM-R. It is a unlabeled dataset of 100+ languages, that is not balanced across languages. We are only interested in the English (\textit{en}) subset, which contained unlabeled strings... % subject to update
    %\item \textbf{AG News Classification}~\cite{zhang2015character} is a dataset for four major classes of news articles, i.e. ``world'',``sports'',``buisiness'',``tech'' with $30$k training and $1.9$k test samples per class. Our motivation for using it is to pretrain BERT~\cite{devlin2018bert} encoders with the proposed ensemble method and then finetune it for news classification.
    %\item \textbf{WikiText-2 Dataset}~\cite{merity2016pointer,merity2016wikitext} is a dataset of more than $100$m tokens from featured articles of wikipedia. It has a vocabulary size of $33278$ with $2088628$, $217646$ and $245569$ tokens taken from $600$, $60$ and $60$ articles from the training, validation and testing sets respectively. The annotation is for the task of token prediction out of context. Our motivation for using it is to pretrain BERT~\cite{devlin2018bert} encoders with the proposed ensemble method and then finetune it for next sentence prediction.
%\end{itemize}
Regarding hardware, we have used multiple A$100{\text -}80$ and A$6000{\text -}48$ cards in PyTorch dataparallel mode~\cite{li2020pytorch}. %The AudioSet and VGGsound preprocessing were done in compute nodes with $32$ hyperthreaded cores and $256$GB of RAM. %The compute configurations were kept the same when training and testing the models on GPU nodes. We have not changed the floating point precision of the CAV-MAE model and we also have not tweaked any hyperparameters except the batch size and learning rate. 

 % \vspace{-2mm}
\subsection{Implementation details}
 % \vspace{-2mm}
%Most of the experiments for trying the various configurations were carried out using the VGGsound dataset due to both its size and versatility. 
%In some the experiments the initial model used is pre-trained on imagenet for both the modalities (this is possible as CAV-MAE treats both video and audio SSL as masked image reconstruction task). For the other variants 
%The AudioSet pre-trained model of CAV-MAE, $\mathsf{scale^{++}}$,  is used as the initial weight. 
For pretraining we have used a batch size of $120$ and for finetuning it is $48$ for VGGsound and 36 for AudioSet20K. The learning rate starts from $0.00005$ and starts decreasing by half every five epochs, after the $15$\textit{th} epoch. For finetuning, the learning rate starts from $0.0001$ and is halved every epoch starting from $2$\textit{nd} epoch. Only a few experiments were done on the much larger AudioSet dataset, due to its sheer size and thus delay in training. In the tables, AudioSet is abbreviated as \textit{AS}, and VGGsound is abbreviated as \textit{VGG}. The metric of measurement is classification accuracy in every form of experiment. \\

% \vspace{-4mm}
\subsection{Experimental results}
 % \vspace{-2mm}
%We have experimented with running the vanilla CAV-MAE model with its original code on the partially available dataset to obtain results on AudioSet and VGGsound.
Table~\ref{tab:vanilla_selfD} shows three rows the first being the results of finetuning on different combinations of both datasets VGG sound and AS-20K on the CAV-MAE. The second one is Moving Average mutual Knowledge Distillation (MAKD)\cite{grill2020bootstrap}, where instead of having shared weights in the two streams the one with lesser loss (teacher) propagated its weight to the one with higher loss (student), for a certain iteration. This role reverses dynamically and thus the \textit{``mutual''} terms come in play. This was our primary attempt towards incorporating KD as the third pillar of SSL after masked data modelling and contrastive learning. The third shows the best result obtained by our proposed ensemble technique of dual complementary mask and self-distillation-based approach.  It can be observed that our ensembled method showed a jump in accuracy over the vanilla technique in the multimodal configuration and the video-only configuration in finetuning. It can also be seen that for AudioSet the proposed method performed better for AV classification. \\

\begin{table}
\scriptsize
\centering
% % \vspace{-3em}
\caption{Comparison of finetuning accuracy (MAKD is Moving Average mutual Knowledge Distillation.)}
\begin{tblr}{
  width = \linewidth,
  colspec = {|Q[240]|Q[130]|Q[140]|Q[142]|Q[160]|Q[170]|Q[170]|},
  hlines,
  %vlines,
}
                  & \tiny{VGG FT AV} & \tiny{VGG FT A only} & \tiny{VGG FT V only} & \tiny{AS-20k FT AV} & \tiny{AS-20k FT A only} &\tiny{AS-20k FT V only} \\
CAV-MAE           & 63.89     & 58.6          & 43.20          & 39.61             & \textbf{37.95}  &    \textbf{31.00}            \\
MAKD &   62.44  & \textbf{59.2} & 43.37 & 40.80 & 37.70 & 30.86 \\
Proposed & \textbf{64.23}     & 58.73         &  \textbf{43.43}        & \textbf{41.03 }          & 37.62 &     30.95

\end{tblr}
% \vspace{-2em}
% % \vspace{-3.5em}
\label{tab:vanilla_selfD}
\end{table}
% \vspace{-4mm}
The final loss propagated with our ensembled method is either determined by dynamic weights as learnable parameters during training or by manually assigning the weights $\lambda_{c}$ and $\lambda_{kd}$ for the contrastive and self-distillation losses respectively. As per CAV-MAE authors, $\lambda_{c}=0.01$ is an optimal choice as per their study so we kept it unchanged and experimented with various values of $\lambda_{kd}$, as shown in Table~\ref{tab:diff_kd_weights}. By overlapped and non-overlapped masks, we mean $n$ random masks $M_i$ so that $M_i \cap M_j \stackrel{}{=} \phi\,\forall \, i,j,i\neq j$ and $M_i \cap M_j = \phi\,\forall \, i,j,i\neq j$, respectively. We observe that for $\lambda_{kd}=10$ with complementary masking, we achieve the best result among all our experiments. Note that in configurations where VGGsound results were not satisfactory, the experiments were not repeated for AudioSet. In this regard, it must be clarified that we have used the abbreviation ``kd'' for self-distillation. Also, the result of adaptive masking strategy~\cite{bandara2023adamae} is compared with our complementary masking strategy. In that work, the mask is adaptively generated based on the input image.
%% % \vspace{-2mm}
\begin{table}[h]
\scriptsize
\centering
\caption{Results of different datasets (for description refer Table~\ref{tab:vanilla_selfD}). The first row shows two random masks with no restrictions on being complementary and $\lambda_{kd}=10$. In the second row onwards, masks are complementary with different values for $\lambda_{kd}$.
}
% % \vspace{-1em}
\begin{tblr}{
  width = \linewidth,
  colspec = {|Q[500]|Q[80]|Q[90]|Q[90]|Q[110]|Q[110]|Q[110]|},
  hlines,
  %vlines,
}
                                       & \tiny{VGG FT AV}      & \tiny{VGG FT A only}  & \tiny{VGG FT V only}  & \tiny{AS-20k FT AV}   & \tiny{AS-20k FT A only} & \tiny{AS-20k FT V only} \\
Adaptive Masking~\cite{bandara2023adamae} & 63.65 & 58.5 & 43.2 & 41.23 & 37.11 & 31.01 \\
With Overlap Dual Mask $\lambda_{kd}$=10    & 63.85          & 58.77          & 43.68          & 41.37 & 37.72 & \textbf{31.59}  \\
No Overlap Dual Mask $\lambda_{kd}$=10 & \textbf{64.23} & 58.73          & 43.43          & 41.03         & 37.62 & 30.95            \\
No Overlap Dual Mask $\lambda_{kd}$=15 & 63.58          & \textbf{58.85} & \textbf{43.71} &  41.13               &   36.57 & 31.12              \\
Proposed No Overlap Dual Mask $\lambda_{kd}$=5  & 64.01          & 58.84          & 43.51          & \textbf{41.62}         &    \textbf{38.42} & 31.25          
\end{tblr}
% \vspace{-2em}
% % \vspace{-3em}
\label{tab:diff_kd_weights}
\end{table}
% % \vspace{-3mm}

The self-distillation strategy is about measuring the divergence between the latent vectors out of the joint encoder and then propagating that as a loss back through the joint encoder, audio and video encoders. 
%The divergence between the outputs of the audio and video encoders was not considered. 
Though while training a multimodal encoder, the joint encoding matters the most, we experimented with the divergences between the outputs of the audio encoders and video encoders of both streams. Our motivation was to study the fact that whether a better joint representation learning can be enhanced by ensuring better modality-specific learning. In Table~\ref{tab:diff_aa_vv} we have the results with the configurations of overlapping dual mask and non-overlapping dual mask in row $1$ and $2$ respectively for the different dataset combinations. We also experimented with the same configurations with dynamic weights replacing $\lambda_{c}$ and $\lambda_{kd}$. In Fig.~\ref{fig:approach}, the violet round denotes the loss mixer, which is named so that the mixing can be either done with manual weights or dynamic weights. The dynamic weights are learnable parameters in the PyTorch computation graph that get updated by the global loss. Row $3$ and $4$ show the results with dynamic weight in Table~\ref{tab:diff_aa_vv}, where row $4$ is the configuration with dynamic weight and divergence on just the joint encoder. It can be observed that dynamic weight delivers better accuracy when distillation between all pairs of encoders (AE, VE, AVE) are considered. %This happens because with so many loss parameters, it is always better to let the model figure out the best way to use them, rather than setting them empirically. 
%% % \vspace{-2mm}

\begin{table} [h]
\scriptsize
\centering
% % \vspace{-1em}
\caption{Results of different datasets (for description refer Table~\ref{tab:vanilla_selfD}) when trained with dual masks and self distillation between the joint encoder (AVE) and also between the audio and video encoders (AE and VE) of the two streams (referred to as AA and VV), respectively. (DW means Dynamic Weight)}
\begin{tblr}{
  width = \linewidth,
  colspec = {|Q[500]|Q[80]|Q[90]|Q[90]|Q[110]|Q[110]|Q[110]|},
  hlines,
  %vlines,
}
                                      & \tiny{VGG FT AV}      & \tiny{VGG FT A only}  & \tiny{VGG FT V only}  & \tiny{AS-20k FT AV}   & \tiny{AS-20k FT A only} & \tiny{AS-20k FT V only} \\
With Overlap Dual Mask  AA VV      & 63.98          & 58.79         & 43.62         & 41.42        & 35.60 & 31.44            \\
No Overlap Dual Mask AA VV    & 64.05          & 58.78         & 43.73        & 41.01        & 37.18  & 30.94           \\
No Overlap Dual Mask AA VV DW & \textbf{64.06} &      \textbf{59.04}         &     43.31       &       \textbf{41.56 }      &      37.57   &  31.58        \\
No Overlap Dual Mask DW       & 63.82          & 58.90          & \textbf{43.83}       &        40.98      &  \textbf{38.28}  &    31.87             
\end{tblr}
% \vspace{-2em}
% % \vspace{-4em}
\label{tab:diff_aa_vv}
\end{table}
 %% \vspace{-2mm}

\begin{table}[h]
\scriptsize
\centering
\caption{Ablation study to demonstrate the effect of multiple student self-distillation  techniques (for description refer Table~\ref{tab:vanilla_selfD}). }
% % \vspace{-2em}
\begin{tblr}{
  width = \linewidth,
  colspec = {|Q[510]|Q[80]|Q[90]|Q[90]|Q[110]|Q[110]|Q[110]|},
  hlines,
  %vlines,
}
                         & \tiny{VGG FT AV} & \tiny{VGG FT A only} & \tiny{VGG FT V only} & \tiny{AS-20k FT AV} & \tiny{AS-20k FT A only} & \tiny{AS-20k FT V only} \\
Triple Mask No Overlap~$\lambda_{kd}=10$   & 63.79     & 58.60          & 43.53         & 40.78           & 38.27 &      31.74      \\
Triple Mask With Overlap~$\lambda_{kd}=10$ & \textbf{64.18}     & \textbf{59.00}           & \textbf{43.85}        & \textbf{41.55}        & \textbf{38.43}    &       30.70  
\end{tblr}
% \vspace{-2em}
% % \vspace{-3em}
\label{tab:triple_stream}
\end{table}

\begin{table}[h]
\scriptsize
\centering
\caption{Finetune accuracy (\%) on different datasets}
% % \vspace{-2em}
\begin{tblr}{
  width = \linewidth,
  colspec = {|Q[6]|Q[4]|Q[4]|Q[4]|},
  hlines,
  % vlines
}
                  & CAV-MAE & Dual Mask With Overlap & Dual Mask Without Overlap \\
AS-2M (V only)    & 30.26           & \textbf{32.59}         & 32.27                     \\
AS-2M (A only)    & 41.71           & 42.27                  & \textbf{42.51}            \\
AS-2M (AV)        & 41.85           & \textbf{43.55}         & 42.73                     \\
AS-20k (V only)   & 28.86           & 30.83                  & \textbf{31.57}            \\
AS-20k (A only)   & 37.21           & \textbf{37.92}         & 37.01                     \\
AS-20k (AV)       & 37.02           & 41.22                  & \textbf{41.34}            \\
ILSVRC (V only)   & 73.73           & 73.57                  & \textbf{73.76}            \\
Kinetics-400 (V only)   & 65.74     & \textbf{66.19}         & 65.78                     \\
Kinetics-400 (AV) & 71.55           & 71.53                  & \textbf{71.69}    
\end{tblr}
% \vspace{-2em}
% % \vspace{-3em}
\label{tab:other_datasets}
\end{table}

\begin{table*}[h!]
\tiny
%\scriptsize
\caption{Results on retrieval, inpainting and visual sound source localization tasks on the VGG-Sound dataset}
\resizebox{\linewidth}{!}{%
\begin{tabular}{|l|crrrrr|cr|c|}
\hline
\multicolumn{1}{|c|}{\multirow{3}{*}{\textbf{Model}}} & \multicolumn{6}{c|}{\textbf{Retrieval}} & \multicolumn{2}{c|}{\multirow{2}{*}{\textbf{Inpainting}}} & \multirow{2}{*}{\textbf{Localization}} \\ \cline{2-7}
\multicolumn{1}{|c|}{} & \multicolumn{3}{c|}{\textbf{Audio → Visual}} & \multicolumn{3}{c|}{\textbf{Visual → Audio}} & \multicolumn{2}{c|}{} &  \\ \cline{2-10} 
\multicolumn{1}{|c|}{} & \multicolumn{1}{c|}{\textbf{R @ 1}} & \multicolumn{1}{c|}{\textbf{R @ 5}} & \multicolumn{1}{c|}{\textbf{R @ 10}} & \multicolumn{1}{c|}{\textbf{R @ 1}} & \multicolumn{1}{c|}{\textbf{R @ 5}} & \multicolumn{1}{c|}{\textbf{R @ 10}} & \multicolumn{1}{c|}{\textbf{Loss A}} & \multicolumn{1}{c|}{\textbf{Loss V}} & \textbf{Avg Cos Sim} \\ \hline
CAV-MAE & \multicolumn{1}{r|}{0.0580} & \multicolumn{1}{r|}{0.1599} & \multicolumn{1}{r|}{0.2142} & \multicolumn{1}{r|}{\textbf{0.0662}} & \multicolumn{1}{r|}{\textbf{0.1917}} & \textbf{0.2537} & \multicolumn{1}{r|}{2.1869} & 2.9981 & \multicolumn{1}{r|}{\textbf{0.2884}} \\ \hline
With Overlap & \multicolumn{1}{r|}{\textbf{0.0594}} & \multicolumn{1}{r|}{\textbf{0.1629}} & \multicolumn{1}{r|}{\textbf{0.2245}} & \multicolumn{1}{r|}{0.0402} & \multicolumn{1}{r|}{0.1367} & 0.1994 & \multicolumn{1}{r|}{2.2948} & 3.0026 & \multicolumn{1}{r|}{0.2807} \\ \hline
No Overlap & \multicolumn{1}{r|}{0.0506} & \multicolumn{1}{r|}{0.1431} & \multicolumn{1}{r|}{0.1944} & \multicolumn{1}{r|}{0.0416} & \multicolumn{1}{r|}{0.1304} & 0.1734 & \multicolumn{1}{r|}{\textbf{1.9985}} & \textbf{2.8756} & \multicolumn{1}{r|}{0.2798} \\ \hline
\end{tabular}}
\label{tab:downstream}
\end{table*}

\begin{table}
\tiny
%\scriptsize
\centering
\caption{Results of different masking strategies on VGG-Sound}
\resizebox{\linewidth}{!}{%
\begin{tabular}{|c|cccc|}
\hline
\multirow{3}{*}{\textbf{Model}} & \multicolumn{4}{c|}{\multirow{2}{*}{\textbf{Classification Accuracy  (\%)}}} \\
 & \multicolumn{4}{c|}{} \\ \cline{2-5} 
 & \multicolumn{1}{c|}{\textbf{Time}} & \multicolumn{1}{c|}{\textbf{Frequency}} & \multicolumn{1}{c|}{\textbf{TF}} & \textbf{Uniform} \\ \hline
CAV-MAE & \multicolumn{1}{c|}{63.58} & \multicolumn{1}{c|}{63.77} & \multicolumn{1}{c|}{63.51} & 63.89 \\ \hline
With Overlap & \multicolumn{1}{c|}{64.91} & \multicolumn{1}{c|}{64.90} & \multicolumn{1}{c|}{64.21} & 63.98 \\ \hline
No Overlap & \multicolumn{1}{c|}{64.83} & \multicolumn{1}{c|}{64.68} & \multicolumn{1}{c|}{64.01} & {64.05} \\ \hline
\end{tabular}}
\label{tab:structured_mask}
\end{table}
 
As discussed in the later part of Section~\ref{sec:combining_losses}, the consideration of more than two parallel streams behaves counter-intuitively as proven by experimental results in Table~\ref{tab:triple_stream}. It can be observed that for three parallel streams, complementary masks perform worse than overlapping masks. The analyzed reason for this behaviour is due to too much data being exposed, especially for the video modality, which overfits the model gradually. For the same reason overlapping masks perform better as the amount of exposed data is bound to be lesser than complementary masks if we take uniform random sampling for keeping unmasked tokens. One interesting observation is that for audioset the triple-stream technique fares better with audio-only finetuning too. The reason is that a larger dataset mitigates overfitting and due to the comprehensive nature of the audio data (spectrogram is for the entire clip, not a single frame sample from a clip of $~300$ frames) the model gets trained better (contradicting videos cannot nullify the knowledge of audio modality). In this regard, it should also be mentioned that the audio-only finetuning fares much better than video-only finetuning as the datasets are annotated for audio and not video. The same can also be observed from the results in Table \ref{tab:other_datasets}, which reports the finetuning accuracy on more datasets like ILSRVC, Kinetics-400 and Audioset 20k subset. Again, it can be seen that classification accuracy only improves when using the proposed dual mask method in all datasets.

Another aspect that has been explored is the evaluation of downstream tasks such as retrieval, localization and inpainting as shown in Table \ref{tab:downstream}. It can be seen that the proposed model performs better in the case of audio-to-visual retrieval and the case of inpainting and is quite competitive for localisation tasks. Yet another important observation can be made by comparing the performance improvement with structured masking vs uniform masking as described in Table~\ref{tab:structured_mask}. It can be concluded that uniform unstructured masking provides the best result when compared to time, frequency or time-frequency(TF) based methods. More results on downstream tasks, masking and visualization are reported in the supplementary material.

In order to demonstrate the efficacy of using self-distillation along with the complementary masking strategy for SSL pretraining, we have fine-tuned various model configurations on the Kinetics dataset
We pre-trained all the encoders (\textit{i.e.} AE, VE and AVE) on Kinetics and we finetuned for human action recognition downstream task. The model was initiated with AudioSet pre-trained weights as the starting point. The results are \textbf{71.90\%, 71.78\% and 71.69\%} for no overlap, with overlap and CAVMAE. We also tested on ILSVRC, with the accuracy of \textbf{73.73\%, 74.53\% and 74.76\%} for CAVMAE, overlap and nonoverlap respectively. These results prove that pretraining with a task-specific dataset makes the encoder learn well (which is obvious!) and self-distillation in some form (complementary masked or overlapping masked input pairs) makes the encoder learn better, unequivocally. It can be observed that the overlapped dual mask performed better than the complementary mask as for Kinetics the annotation is based on video rather than audio, thus overexposure of video modality caused it to overfit. %write something after ImageNet results comes \\
%% \vspace{mm}

From the experiments conducted, we can summarize our findings about our proposed model in the following points: complementary masks have performed well or generated competitive results depending on the downstream task, especially on multimodal datasets. As the video sampling is not as uniform as audio, complementary masks don't scale well when the downstream annotation is based on video instead. However, for audio-based annotation in downstream tasks, complementary masking scales up the results elegantly. Moreover, the induction of knowledge distillation with KL-divergence on the complementary brings in the modular correspondence. This makes the encoder struggle to learn mask agnostic modelling and the model starts to learn fine modality-specific information. Further, along with the mask encoder that forces encoding the input information in the fusion and the constructive task that helps to find the explicit audio-visual correspondence, the knowledge-distillation helps to find the modular correspondence as results provide better modelling.
%\end{itemize}

%% % \vspace{-2mm}

% % \vspace{-1em}
\section{Conclusion}
% \vspace{-1em}
In the quest for finding an optimal learning strategy for SSL on large datasets, we ended up exploring many untrodden nooks and corners of knowledge distillation and masked data modelling. Masked autoencoders first leveraged the idea of masked data modelling on image modality with reconstruction from incomplete data as the sole learning objective. Later CAV-MAE elevated that for training a task-agnostic multimodal learner by incorporating intermodal contrastive learning as an extra learning objective. In our proposed learning strategy we have demonstrated that adding knowledge distillation, driven by divergence between embeddings from two inputs subjected to complementary masks, elevated the result still further. We have compared our results with the state-of-the-art methodology for the task, CAV-MAE and found that the proposed KDC-MAE achieved better results. This work opened the door for many paths to explore and future versions for more extensive studies.
%with the existing and other novel learning objectives for SSL. The quest to find the holy grail of SSL - building a task, modality and input size agnostic generic learner, will live on. 

%%%%%%%%% REFERENCES
{\small
\bibliographystyle{ieee_fullname}
\bibliography{egbib}
}

\end{document}